\documentclass[journal]{IEEEtran}

\usepackage{amsmath,amssymb,amsfonts}

\usepackage[english]{babel}
\usepackage{algorithm,algorithmic}
\usepackage{graphicx}
\usepackage{textcomp}
\usepackage{color}
\usepackage{lipsum}
\usepackage{epstopdf}
\usepackage{amsopn}
\usepackage{amsthm}
\usepackage{enumerate}
\usepackage{cancel}
\usepackage{mathrsfs}
\usepackage{mathdots}
\usepackage{euscript}
\usepackage{amscd}
\usepackage{cite}
\usepackage{placeins}
\usepackage[latin1]{inputenc}
\usepackage{tikz}
\usetikzlibrary{snakes,arrows,shapes}

\usepackage{centernot}
\usepackage{subcaption}

\usepackage{threeparttable}
\usepackage{tabularx}
\usepackage{booktabs}

\newtheorem{theorem}{Theorem}

\newtheorem{lemma}{Lemma}
\newtheorem{proposition}{Proposition}

\theoremstyle{definition}
\newtheorem{definition}{Definition}

\theoremstyle{remark}\newtheorem{remark}{Remark}
\newcommand{\Bfk}{\mathfrak{B}}
\newcommand{\Bfkp}{\mathfrak{B}_+}
\newcommand{\Bfko}{\mathfrak{B}_o}
\newcommand{\Bfkop}{\mathfrak{B}_{o,+}}
\newcommand{\tr}{\mathrm{tr}}
\newcommand{\al}[1]{\begin{align}#1\end{align}}
\newcommand{\bmat}{\begin{bmatrix}}
\newcommand{\emat}{\end{bmatrix}}

\newcommand{\Ls}{\mathbb{L}}
\newcommand{\Rs}{\mathbb{R}}
\newcommand{\Ts}{\mathbb{T}}

\newcommand{\Ltwomat}{\Ls^2_{m\times m}(\Ts)}
\newcommand{\Lonemat}{\Ls^1_{m\times m}(\Ts)}

\DeclareMathOperator{\diag}{diag}
\DeclareMathOperator{\trace}{tr}

\newcommand{\E}{{\mathbb E}}
\newcommand{\Rbb}{\mathbb R}

\newcommand{\Cbb}{\mathbb C}

\newcommand{\Zbb}{\mathbb Z}

\newcommand{\Nbb}{\mathbb N}
\newcommand{\Tbb}{\mathbb T}

\newcommand{\Db}{\mathbf D}

\newcommand{\Hb}{\mathbf H}

\newcommand{\Pb}{\mathbf P}
\newcommand{\Qb}{\mathbf Q}

\newcommand{\Scal}{\mathcal{S}}

\renewcommand{\d}{\mathrm{d}}

\renewcommand{\Re}{\mathrm{Re}}

\newcommand{\F}{\mathrm{F}}

\newcommand{\nn}{\nonumber}

\begin{document}

\title{On the Spectral Estimation of VARMA Processes}

\author{Bin Zhu and Mattia Zorzi
\thanks{B.~Zhu is with the School of Intelligent Systems Engineering, Sun Yat-sen University, Gongchang Road 66, 518107 Shenzhen, China (email: \texttt{zhub26@mail.sysu.edu.cn}).}
\thanks{M.~Zorzi is with the Department of Information Engineering, University of Padova, Via Giovanni Gradenigo, 6b, 35131 Padova, Italy (email: \texttt{zorzimat@dei.unipd.it}).}
\thanks{This work was partially supported by:  i) the National Natural Science Foundation of China under the grant number 62103453; ii)  the ``Hundred-Talent Program'' of Sun Yat-sen University; iii) the Italian Ministry of University and Research under the PRIN project ``Extracting essential information and dynamics from complex networks'', grant no. 2022MBC2EZ. Corresponding author B.~Zhu. Tel. +86 14748797525. Fax +86(xx) xxxxxxxx.}%
}

\maketitle

\begin{abstract}
	We consider the problem of estimating the power spectral density of a Vector-valued Autoregressive Moving-Average (VARMA) process. A well-established approach to accomplishing this task consists of formulating a moment problem that selects the spectral density maximizing an entropy-like index while matching certain moments of the underlying process. However, existing methods assume that the moving-average (MA) component is either fixed or scalar, thereby reducing the class of ARMA models over which the optimal spectral density is sought. In this paper, we propose a new moment problem, based on the notion of transportation entropy, for learning the spectral density of a VARMA process without fixing the MA component or restricting it to be scalar.
 \end{abstract}


\section{Introduction}\label{sec:intro}

Autoregressive Moving-Average (ARMA) {processes} have wide applications in different domains, including systems and control \cite{10588982,Cao-Li-2025-ARMAX}, signal processing \cite{stoica2005spectral}, time series analysis \cite{brillinger2001time}, and finance \cite{tsay2005analysis}.
From the perspective of identification and control, these processes have rational spectra which can approximate any continuous spectral density if the order of the model is sufficiently large. Moreover, standard realization theory can convert the ARMA model into a state-space form which is convenient, e.g., for the purpose of filtering.

{It is often the case that the spectral density of ARMA processes must be identified from data.}
For scalar processes, a well-posed problem of jointly estimating the AR and MA parameters was formulated in \cite{byrnes2001cepstral,enqvist2004aconvex} as maximizing the entropy of the process with equality constraints of covariance and cepstral matching (i.e., the moment equations). The sample version of the latter can readily be obtained from a finite-length realization of the underlying process. It was shown that the correspondence between the covariances and the AR parameters is a smooth bijection, and the same holds for the correspondence between the cepstral coefficients and the MA parameters. The cepstral coefficients are also called \emph{logarithmic moments} because they are the Fourier coefficients of $\log \Phi$ where $\Phi$ (assumed to be positive) {denotes} the spectral density of the process.

For vector-valued ARMA (VARMA) processes, the model parameters become matrices, and the estimation problem becomes more challenging. A number of attempts have been made to extend the Maximum Entropy framework to the multivariate case. The first setup considered the case where the MA part is fixed \emph{a priori}, and certain divergence indices are minimized subject to matricial covariance constraints; see, e.g., \cite{FPR-08,FMP-12,Z-14}.
However, if we want to jointly estimate the matricial AR and MA parameters from the data, the standard Maximum Entropy method \cite{burg1975maximum} works only with \emph{scalar logarithmic moments} \cite{lindquist2013onthemultivariate,lindquist2016modeling,10588982}, bringing about VARMA models with scalar MA parts, i.e., the MA parameters have the form of a scalar times the identity matrix. We remark that the scalar logarithmic moments of a {vector-valued} process with a matricial spectral density $\Phi$ (assumed to be positive definite) were first defined in \cite{BLN-03} as the Fourier coefficients of $\log\det\Phi$. These quantities were utilized in the spectral estimation of VARMA process in later works that are cited above. {It is worth noting that assuming the MA component to be fixed or scalar reduces the class of ARMA models over which the optimal spectral density is sought, potentially resulting in a spectral estimator that is not always satisfactory.}

One may be tempted to use the matricial logarithm for positive definite matrices to define a kind of matricial cepstral coefficients and hope to generalize the multivariate spectral estimation method with the scalar logarithmic moments.
{As we will see}, however, it is not possible to go through the Lagrange dual analysis of the corresponding Maximum Entropy problem, because the optimal primal variable (which is a function) cannot be expressed in terms of the dual variable. 
Notice that a successful dual analysis is the key to obtaining  a finite-dimensional dual problem, which serves as a practical tool for estimating the parameters characterizing the spectral density.
The same happens if we consider the more general notion of $\nu$-entropy (parametrized by a positive integer $\nu$) proposed in \cite{Zhu-Zorzi2023cepstral,Zhu-Zorzi-2022-cep-est} for the scalar case.

Hence, {the} major obstacle in estimating {a VARMA process by means of a moment problem}  is to find a suitable definition of the cepstral coefficients for a vector-valued process, and this is precisely the motivation behind the current work. Inspired by recent results on optimal transportation between Gaussian processes \cite{Z-MED,Z-TRANSP,FPR-08}, we define a new concept of matricial cepstral coefficients using a spectral factor $W$ of the spectral density of the process. For this reason, they are termed ``$W$-cepstral coefficients''. These coefficients are naturally compatible with the transportation entropy derived from the transportation (Hellinger) distance between two spectral densities. We show that this new Maximum Entropy formulation is amenable to Lagrange dual analysis, and the dual problem becomes well-posed after incorporating an appropriate design of regularization. Then we derive a spectral estimator, based on the proposed Maximum Entropy problem,  which uses only a finite sample path of {the underlying} process. {The advantage of our VARMA spectral estimator} is illustrated by numerical experiments.

The rest of the paper is organized as follows.
Sec.~\ref{sec:spec_approx} {introduces the VARMA spectral estimation problem and two possible Maximum Entropy formulations}, and explains the difficulty.
Sec.~\ref{sec:OT_formul} introduces the transportation entropy and the $W$-cepstral coefficients and formulates {the} new Maximum Entropy problem. A detailed dual analysis is carried out.
Sec.~\ref{sec:reg_dual_prob} regularizes the dual optimization problem and investigates its well-posedness property.
Sec.~\ref{sec:est} provides a spectral estimation procedure based on the regularized dual problem and a finite sample path of a random process, with a discussion of the statistical consistency of the estimator.
Sec.~\ref{sec:sims} illustrates our VARMA spectral estimator in comparison with some alternative ones through numerical simulations.
Sec.~\ref{sec:conclus} concludes the paper.

{\em Notation.} Let $\Tbb:=[0, 2\pi)$ be the interval for normalized angular frequencies which is isomorphic to the unit circle $\{z\in\Cbb : |z|=1\}=\{e^{i\vartheta} : \vartheta\in\Tbb\}$ in the complex plane. 
The integral of a function $f$ on $\Tbb$ with respect to the normalized Lebesgue measure, namely $\int_{\Tbb} f(\vartheta) \frac{\d\vartheta}{2\pi}$, is often abbreviated as $\int f$. The symbol $\Ls^p_{m\times n}(\Ts)$ with $0<p\leq \infty$ represents the vector space of $m\times n$ complex matrix-valued functions on $\Ts$ whose entries are complex-valued functions in $\Ls^p(\Ts)$. The set ${\cal S}_{m,+}$ consists of  (Hermitian) positive semidefinite matrix-valued functions in $\Lonemat$.

\section{VARMA Spectrum approximation}\label{sec:spec_approx}

We consider the problem of finding a power spectral density $\Phi\in{\cal S}_{ m,+}$ describing a {vector-valued ARMA Gaussian}  process assuming that some moments of its spectrum are known. In the univariate case, i.e., $m=1$, a solution can be obtained from the maximum entropy problem \cite{enqvist2004aconvex}:
\al{\label{ME_pb}& \underset{\Phi\in{\cal S}_{m,+}}{ \max\;}  \int\log\det  \Phi\nn\\ & \hbox{ s.t. }  \int e^{i\vartheta k}  \Phi=R_k,\quad k=0, \ldots, n\nn \\
 &\hspace{0.5cm}   \int e^{i\vartheta k}   \log \Phi=C_k,\quad k=1, \ldots, n  }
where the objective function denotes the entropy rate \cite{cover2006elements} of the Gaussian process with a spectrum $\Phi$, and in the moment constraints $R_k\in\Rs^{m\times m}$, with $k=0\ldots n$, and {$C_k\in \Rs^{m\times m}$}, with $k=1\ldots n$, are the covariance lags and the cepstral coefficients of the spectrum, respectively.

In the multivariate case, i.e., $m>1$, we can still consider \eqref{ME_pb} where $\log \Phi$ denotes the matricial logarithm which is well defined when $\Phi$ is positive definite pointwise for each $\vartheta\in\Tbb$. In that case, $\Phi$ admits a spectral decomposition $\Phi=X\Lambda X^*$ in which $X$ is a unitary matrix of eigenvectors and $\Lambda$ is a diagonal matrix of positive real eigenvalues, both of which are functions of $\vartheta$. Then $\log \Phi := X(\log \Lambda) X^*$ where $\log \Lambda$ is a diagonal matrix obtained by taking the logarithm of each diagonal entry of $\Lambda$.
{As a first attempt, we try to solve Problem \eqref{ME_pb}.} Let $\mathbf Q$ be an element of the vector space
\begin{equation}
	\Bfk := \left\{ \{Q_k\}_{k=0}^n : Q_k\in\mathbb R^{m\times m}\ \text{and}\ Q_0^T=Q_0 \right\}
\end{equation}
and $\mathbf P$  {an element}  of
\begin{equation}
	\Bfko:=\left\{\{P_k\}_{k=1}^n : P_k\in\mathbb R^{m\times m}\right\}.
\end{equation}
The Lagrangian of \eqref{ME_pb} is 
 \al{L(\Phi, &\,\Pb, \Qb)=   \int \log \det\Phi- \tr \left[\sum_{k=0}^n Q_k^T   \left(\int e^{i\vartheta k}  \Phi -R_k\right)\right]\nn\\
 &+\tr\left[\sum_{\substack{k=1}}^n P_k^T\left(\int e^{i\vartheta k}   \log \Phi-C_k\right)\right]\nn \\
=&  \tr \left(\int P  \log \Phi-Q\Phi\right) +\tr\sum_{k=0}^n Q_k^{T} R_k-\tr\sum_{\substack{k=1}}^n P_k^{T} C_k \nn  }
where 
	\begin{equation}\label{def_poly_P_Q}
		\begin{aligned}
			Q(e^{i\vartheta}):=Q_0+\frac{1}{2}\sum_{k=1}^n \left( Q_k e^{-i \vartheta k}+Q_k^T e^{i\vartheta k}\right) \\
				\text{and}\  
			  P(e^{i\vartheta}):=I+\frac{1}{2}\sum_{k=1}^n \left( P_k e^{-i\vartheta k}+P_k^T e^{i\vartheta k} \right)
		\end{aligned}
	\end{equation}
	are polynomial matrices constructed from $\Qb$ and $\Pb$, respectively. Under the assumption that $P$ is positive definite on $\Ts$,
it is not difficult to see that $L(\cdot, \Pb, \Qb)$ is concave, and the arguments are similar to those in \cite{RKL-16multidimensional}. 
 Then the unconstrained maximizer is also a stationary point, i.e., it is obtained by setting equal to zero the first variation of $L$ along the direction $\delta \Phi\in \Lonemat$:
 \al{\delta L &(\Phi,\Pb, \Qb;\delta\Phi)= \tr  \int \left[  P \int_0^\infty (\Phi +t I)^{-1}  \right.\nn\\ 
& \left.\times \delta\Phi (\Phi +t I)^{-1} \mathrm d t -Q \delta \Phi\right]\nn
 }
 where we exploited \cite[Proposition A.2]{Z-14}. Accordingly, the stationary point satisfies the condition
   \al{ \label{stat_log} \int_0^\infty (\Phi +t I)^{-1}  P   (\Phi +t I)^{-1}\mathrm d t \mathrm  =Q. } 
It is clear that it is not possible to find a closed-form expression for $\Phi$ in terms of the Lagrange multipliers, and thus we cannot carry out the dual analysis.
 
{\begin{remark} \label{RemSCAL} If, in Problem \eqref{ME_pb}, we replace $\log \Phi$ with $\log \det \Phi$ (and thus also $C_k\in \Rs^{m\times m}$ with $c_k\in\Rs$), i.e., we consider the scalar logarithmic moments of a {vector-valued} process defined in \cite{BLN-03}, then the matricial polynomial $P$ defined in \eqref{def_poly_P_Q} is replaced with a scalar trigonometric polynomial \al{\label{scal_pol}p(e^{i\vartheta})=1+\sum_{k=1}^n p_k \cos(\vartheta k),\quad p_k\in\Rs} and the stationarity condition in \eqref{stat_log} reduces to
\al{ p\Phi^{-1}=Q.\nn}
Moreover, it is possible to show, through the dual problem, that the solution to the primal problem exists and is of the form \cite{10588982}:
\al{\label{optSpec_scal} \Phi^\circ= p Q^{-1}.}
However, the spectrum in \eqref{optSpec_scal} describes a special class of ARMA processes. Indeed, let $b(z)=\sum_{k=0}^n b_k z^{-k}$, $b_k\in\Rs$, and $A(z)=\sum_{k=0}^n A_k z^{-k}$, $A_k\in\Rs^{m\times m}$, be the minimum phase spectral factors of $p$ and $Q$, i.e., $p(z)=b(z^{-1})b(z)$ and $Q(z)=A(z^{-1})^TA(z)$.
Then the process $\{y_t\}_{t\in \Zbb}$ with a spectrum \eqref{optSpec_scal} admits the \emph{unilateral} ARMA representation 
\al{\label{modellSS}\sum_{k=0}^k A_k y_{t-k}=\sum_{k=0}^n b_k e_{t-k} }
where  $\{e_t\}_{t\in \Zbb}$ is a normalized white Gaussian noise. We conclude that such a process is very special, since its MA component is scalar, i.e., the coefficients $b_k$ are scalar.
 \end{remark}}

Another possible route is to consider a general definition of entropy, called \emph{$\nu$-entropy}, which is given in \cite{Zhu-Zorzi2023cepstral,Zhu-Zorzi-2022-cep-est} for the scalar case, i.e. $m=1$, with an integer parameter $\nu\geq 1$. In the case of $\nu=1$, it reduces to the classic entropy in \eqref{ME_pb}. Its extension to the matricial case ($m>1$) with $\nu\geq 2$ is 
\al{\frac{\nu^2}{\nu-1}\tr \int  \Phi^{\frac{\nu-1}{\nu}}\nn.} 
 Now in the moment problem, we can replace the entropy rate with the $\nu$-entropy above, and the cepstral coefficients with the matricial $\nu$-cespral coefficients, and obtain
 \al{\label{MEnu_pb}& \underset{\Phi\in\Scal_{m,+}}{ \max\;}\frac{\nu^2}{\nu-1}\tr \int  \Phi^{\frac{\nu-1}{\nu}}\nn\\ & \hbox{ s.t. }  \int e^{i\vartheta k}  \Phi=R_k,\quad k=0, \ldots, n\nn \\
 &\hspace{0.5cm} \frac{\nu^2}{\nu-1} \int e^{i\vartheta k}   \Phi^{\frac{\nu-1}{\nu}}=C_k,\quad k=1, \ldots, n. }
The notion of matricial $\nu$-cespral coefficient is an extension of the one given for scalar processes ($m=1$) in \cite{Zhu-Zorzi2023cepstral}. More precisely,
the matricial power function with an exponent $\alpha>0$ is defined in a way similar to the matricial logarithm. In fact, we let $\Phi^\alpha := X \Lambda^{\alpha} X^*$ where $X$ and $\Lambda$ come from the (pointwise) spectral decomposition $\Phi=X\Lambda X^*$ and the power function $(\cdot)^\alpha$ is applied to each (nonnegative) eigenvalue on the diagonal of $\Lambda$.  The Lagrangian of \eqref{MEnu_pb} is 
 \al{L(\Phi,&\, \Pb, \Qb)=  \frac{\nu^2}{\nu-1}\tr \int \Phi^{\frac{\nu-1}{\nu}}\nn\\ 
 & - \tr \left[\sum_{k=0}^n Q_k^T   \left(\int e^{i\vartheta k}  \Phi -R_k\right)\right]\nn\\
 &+\tr\left[\sum_{\substack{k=1}}^n P_k^T\left(\frac{\nu^2}{\nu-1}\int e^{i\vartheta k}   \Phi^{\frac{\nu-1}{\nu}}-C_k\right)\right]\nn \\
=&  \tr \!\left(\int \!\!\frac{\nu^2}{\nu-1}  P  \Phi^{\frac{\nu-1}{\nu}} \!-\! Q\Phi \!\right)\! +\!\tr\sum_{k=0}^n Q_k^{T} R_k \!-\!\tr\sum_{\substack{k=1}}^n P_k^{T} C_k \nn  }
where $\Pb\in \Bfko$ and $\Qb\in\Bfk$. {We associate the polynomial matrices $P$ and $Q$, defined in \eqref{def_poly_P_Q}, to $\Pb$ and $\Qb$, respectively.} Under the assumption that $P$ is positive definite on $\Ts$, it can be shown that $L(\cdot, \Pb, \Qb)$ is concave \cite{Z-14,Z-14rat}.
Then, the unconstrained maximizer is also a stationary point, i.e., it is obtained by setting equal to zero the first variation of $L$ along the direction $\delta \Phi\in\Lonemat$:
 \al{\delta L &(\Phi,\Pb, \Qb;\delta\Phi)= \tr  \int \left[\nu P\int_0^1\int_0^\infty \Phi^{\frac{\nu-1}{\nu}(1-\tau) }(\Phi +t I)^{-1}  \right.\nn\\ 
& \left.\times \delta\Phi (\Phi +t I)^{-1}\Phi^{\frac{\nu-1}{\nu}  \tau  } \mathrm d t \mathrm d\tau -Q \delta \Phi\right]\nn
 } where we exploited \cite[Proposition A.3]{Z-14}. Accordingly, the stationary point satisfies the condition
 \al{\nu\! \int_0^1 \!\int_0^\infty \!(\Phi +t I)^{-1}  \Phi^{\frac{\nu-1}{\nu}\tau }P \Phi^{\frac{\nu-1}{\nu}(1-\tau) } (\Phi +t I)^{-1}\mathrm d t \mathrm d\tau \!= \! Q. \nn} 
 Also in this case it is not possible to find a closed-form expression for $\Phi$ in terms of the Lagrange multipliers and thus we cannot continue the dual analysis.
 
\begin{remark} One could replace the matricial $\nu$-cespral coefficient with its scalar version for vector-valued processes, i.e., we replace $ \Phi^{\frac{\nu-1}{\nu}}$ with $ \tr[\Phi^{\frac{\nu-1}{\nu}}]$ (and thus also $C_k\in \Rs^{m\times m}$ with $c_k\in\Rs$). The Lagrangian in this case reduces to
 \al{ \tr \!\left(\int \!\!\frac{\nu^2}{\nu-1}  p  \Phi^{\frac{\nu-1}{\nu}} \!-\! Q\Phi \!\right)\! +\!\tr\sum_{k=0}^n Q_k^{T} R_k \!-\!\tr\sum_{\substack{k=1}}^n p_k c_k \nn  }
where  the scalar polynomial $p$ is defined as in \eqref{scal_pol}. Under the assumption that $p$ is positive on $\Ts$, it is not difficult to prove that the Lagrangian above is concave with respect to $\Phi$ and thus its unconstrained maximum is obtained by imposing the stationarity condition, which turns out to be  \al{\int (\nu  p  \Phi^{-\frac{1}{\nu} } -Q)\delta \Phi=0, \quad \forall \; \delta \Phi\in\Lonemat. \nn} 
Under the assumption that $Q$ is positive definite on $\Ts$ we have the optimal form of the primal variable 
\al{\label{optSpec_scal2}\Phi^\circ =\nu^\nu p^\nu Q^{-\nu}. }
Thus, the spectrum in \eqref{optSpec_scal2} describes a special class of ARMA processes
which admit a time-domain representation  similar to the one in \eqref{modellSS}.
\end{remark}
  
\section{Optimal transport formulation}\label{sec:OT_formul}

It is worth noting that an equivalent formulation of problem  (\ref{ME_pb})
is to minimize $\Db_{\text{IS}}(\Phi,I)$ under the moment constraints and 
\al{\Db_{\text{IS}}(\Phi,\Psi)=\int -\log\det \Phi+\log\det \Psi +\tr(\Phi\Psi^{-1})-{m}}
denotes the Itakura--Saito distance between the spectral densities $\Phi$ and $\Psi$, \cite{FMP-12}. Indeed,
\al{\Db_{\text{IS}}(\Phi,I)=\int -\log\det \Phi+\tr(\Phi)-{m},}
and the last two terms are constant over the feasible set characterized by the moment constraints. Similarly, an equivalent formulation of Problem  (\ref{MEnu_pb})
is to minimize $\Db_{\beta}(\Phi,I)$, with $\beta=1-\nu^{-1}$, under the moment constraints  and 
\al{\Db_{\beta}(\Phi,\Psi)=\tr\int \frac{1}{\beta-1}(\Phi^\beta -\Phi\Psi^{\beta-1})-\frac{1}{\beta}(\Phi^\beta-\Psi^\beta)\nn} denotes the $\beta$-divergence between the spectral densities $\Phi$ and $\Psi$ \cite{Z-14}.
Drawing inspiration from that, we set up the moment problem using the optimal transportation between Gaussian stationary processes 
 \cite{Z-TRANSP} that has been previously introduced in
\cite{FPR-08}:
 \al{\Db_{\text{TD}}(\Phi,\Psi)^{2}=&\min_{\substack{W_\Phi,W_\Psi\in \Ls^2_{m\times m}(\Ts)}}\tr\int (W_\Phi-W_\Psi)(W_\Phi-W_\Psi)^*\nn\\
 & \hbox{s.t. } W_\Phi W_\Phi^*=\Phi \nn\\ 
 &\hspace{.5cm}W_\Psi W_\Psi^*=\Psi \label{transport_dist}
}
 where $W_\Phi$ and $W_\Psi$ represent a left spectral factor of $\Phi$ and $\Psi$, respectively.
Then we define the transportation entropy as
 \al{\Hb_{\text{T}}(\Phi) & := -\left(\Db_{\text{TD}}(\Phi,I)^2 - \trace\int\Phi -m \right) \nn \\
 	& = \max_{\substack{W\in\Ltwomat }} \tr\int W+W^*\nn\\ 
 &\hspace{.5cm}\hbox{ s.t. } W W^*=\Phi. \label{def_transport_entropy}}
Notice that the term $-\trace\int\Phi- m$ can be treated as a constant in the constrained maximization problem above, and we added it only to simplify the subsequent analysis. Indeed, if we define another functional $\tilde\Hb_{\text{T}}(\Phi):=-\Db_{\text{TD}}(\Phi,I)^2$, then for each fixed $\Phi$, the corresponding maximization problem will admit an optimal spectral factor $W^\circ$ that is identical to the maximizer of \eqref{def_transport_entropy}. The maximum values of the two objective functionals, however, will clearly be different.

Before formulating the problem, we need to define the cepstral coefficients. 
 \begin{definition}\label{def:W_ceps}
 	Let  $W$ be a left spectral factor of the spectral density $\Phi$. We define as the $k$-th matricial $W$-cepstral coefficient of $\Phi$ the matrix
 \al{C_k=\frac{1}{2}\int e^{i\vartheta k} (W+W^*), \quad k\in\Zbb.} 
 \end{definition}
 
It is clear that the definition above depends on the particular spectral factor. If we take $W=\Phi^{1/2}$, then we have 
  \al{\label{mat_cep_sqrt_factor}C_k= \int e^{i\vartheta k}\Phi^{1/2}} 
  which coincides (up to a constant factor) with the $\nu$-cepstral coefficient in \eqref{MEnu_pb} with $\nu=2$.

Thus, we obtain the moment problem:
 \al{\label{pb_primal}& \underset{\substack{\Phi\in\Lonemat,\\ W\in\Ltwomat}}{\max\;} \Hb_{\text{T}}(\Phi)\nn\\ & \hbox{ s.t. }  \Phi = W W^* \nn \\
 	& \hspace{0.5cm} \int e^{i\vartheta k}  \Phi=R_k,\quad k=0, \ldots, n\nn \\
 &\hspace{0.5cm} \frac{1}{2} \int e^{i\vartheta k}   (W+W^*)=C_k,\quad k=1, \ldots, n. }
Due to the presence of the constraint $\int\Phi=R_0$ with $R_0$ fixed, the term $\trace\int\Phi$ in \eqref{def_transport_entropy} becomes a {constant} for the optimization problem, and we can use the objective functional in the maximization problem of \eqref{def_transport_entropy}.
In fact, the optimization problem \eqref{pb_primal} can be rewritten in terms of $W$ as
 \al{& \underset{W\in\Ltwomat}{ \max\;}\tr\int W+W^*\nn\\ & \hbox{ s.t. }  \int e^{i\vartheta k}  WW^*=R_k,\quad k=0, \ldots, n\nn \\
 &\hspace{0.5cm} \frac{1}{2} \int e^{i\vartheta k}   (W+W^*)=C_k,\quad k= 1, \ldots, n. }
The Lagrangian of the above problem is \al{L(W,&\,\Pb, \Qb)=  \tr \int W+W^*\nn\\ & - \tr\left[\sum_{k=0}^n Q_k^T   \left(\int e^{i\vartheta k}  WW^* -R_k\right)\right]\nn\\
 & +\tr\left[\sum_{\substack{k=1}}^n P_k^T\left( \int e^{i\vartheta k}   (W+W^*)-2C_k\right)\right]\nn \\
& =  \tr \int \left[ (W+W^*)P-WW^*Q\right] \nn \\ 
& \hspace{.5cm} +\tr\sum_{k=0}^n Q_k^{T} R_k-2\tr\sum_{\substack{k=1}}^n P_k^{T} C_k, \label{Lagrangian}}  
where $(\Pb,\Qb)\in\Bfko\times \Bfk$ are  the Lagrange multipliers. 
Next, we consider the unconstrained maximization  problem 
\al{\label{sup_Lagrangian}\underset{W\in\Ltwomat}{\sup} \; L(W,\Pb,\Qb).}
We first show that in order for the supremum above to be finite, the polynomial matrix $Q$ must be positive semidefinite everywhere on $\Tbb$. To this end, let us suppose the contrary, i.e., there exist $\bar \vartheta\in\Ts$ and a vector $v\in\mathbb C^m$ such that $\|v\|=1$ and $v^* Q(e^{i\bar\vartheta}) v<0$. Then, according to continuity, there exists a neighborhood of $\mathcal B$ centered in $\bar \vartheta$ such that 
\al{v^* Q(e^{i\vartheta}) v<0 \quad \forall \, \vartheta\in\mathcal B.} 
We take the Hermitian spectral factor 
\al{W_\alpha(e^{i\vartheta}) := \left\{\begin{array}{ll}\alpha \left|v^* Q(e^{i\vartheta}) v\right|vv^*, & \vartheta\in\mathcal B \\
O,  & \hbox{otherwise} \end{array}\right.}
with $\alpha\in\mathbb R$. Plugging $W_\alpha$ into the Lagrangian \eqref{Lagrangian}, we have
\al{L&(W_\alpha,\Pb,\Qb)=\tr \int_{\mathcal B} \left[ (W_\alpha+W_\alpha^*)P-W_\alpha W_\alpha^*Q \right] +\kappa\nn\\
& = \int_{\mathcal B} \{ 2\alpha v^* P(e^{i\vartheta}) v\left |v^* Q(e^{i\vartheta}) v \right|- \alpha^2\underbrace{[v^* Q(e^{i\vartheta}) v]^3}_{<0} \} +\kappa\nn}
where $\kappa$ denotes the terms that do not depend on $W_\alpha$. Thus, $L(W_\alpha,\Pb,\Qb)\rightarrow \infty$ as $|\alpha|\rightarrow \infty$. Therefore, the supremum in \eqref{sup_Lagrangian} is finite only if $\Qb\in\overline{\Bfkp}$ where 
\al{\label{Bp}\Bfkp:=\{\Qb\in\Bfk : Q>0 \hbox{ on } \Ts\}.} 

Next, we handle the singularity of $Q$. More precisely, let us assume that $Q$ is \emph{always} singular on $\Tbb$. In other words, the polynomial $\det Q(e^{i\vartheta})\equiv 0$ for any $\vartheta\in\Tbb$. 
Suppose that there exists $\bar \vartheta\in\Ts$ such that $Q(e^{i\bar\vartheta})$ has a normalized eigenvector $\bar v$ associated to the zero eigenvalue and $\bar v^* P(e^{i\bar \vartheta}) \bar v\neq 0$. According to \cite[Thm.~6.3.12, p.~409]{horn2013matrix}, the eigenvector of a polynomial matrix $Q(e^{i\vartheta})$ associated with the \emph{smallest} eigenvalue $\lambda=0$ can be chosen continuously; see also \cite{kato2012perturbation,Lax2013LA}. Hence $Q(e^{i\vartheta})$ admits an eigenvector function $v(\vartheta)$ associated with the zero eigenvalue that satisfies $v(\bar \vartheta)=\bar v$.
By continuity, there exists a neighborhood $\mathcal B$ centered in $\bar\vartheta$ such that 
\al{v(\vartheta)^* P(e^{i\vartheta}) v(\vartheta)\neq 0 \quad \forall \, \vartheta\in\mathcal B.\nn} 
We can now take
\al{W_\alpha(e^{i\vartheta}) := \left\{\begin{array}{ll}\alpha  [v(\vartheta)^* P(e^{i\vartheta}) v(\vartheta)] v(\vartheta)v(\vartheta)^*, & \vartheta\in\mathcal B, \\0 ,  & \hbox{otherwise.} \end{array}\right.}
Then, we have 
\al{L&(W_\alpha,\Pb,\Qb)=\tr \int_{\mathcal B} \left[ (W_\alpha+W_\alpha^*)P-W_\alpha W_\alpha^*Q \right]+\kappa\nn\\
&=\int_{\mathcal B} 2\alpha \left[v(\vartheta)^* P(e^{i\vartheta}) v(\vartheta) \right]^2 +\kappa\rightarrow \infty\nn}
as $\alpha \to \infty$. The second equality above holds because $v(\vartheta)^* Q(e^{i\vartheta}) v(\vartheta)\equiv0$. We conclude that the supremum of the Lagrangian is finite only if
\begin{equation}\label{cond2}
v\in\ker Q(e^{i\vartheta}) \implies v^*P(e^{i\vartheta})v=0\quad 	\forall\vartheta\in\Tbb.
\end{equation}
If $P$ is also positive semidefinite (or negative semidefinite) pointwise, then the condition \eqref{cond2} boils down to
\begin{equation}\label{cond3}
	\ker Q(e^{i\vartheta}) \subseteq \ker P(e^{i\vartheta})\quad \forall\vartheta\in\Tbb,
\end{equation}
or {equivalently},
\al{\mathrm{Range}\,P(e^{i\vartheta})\subseteq \mathrm{Range}\,Q(e^{i\vartheta})  \quad \forall \, \vartheta\in\Ts. \label{cond4}}
In general, the condition \eqref{cond3} or \eqref{cond4} implies \eqref{cond2} but the converse may not hold when $P$ is indefinite.
Under conditions $\Qb\in\overline{\Bfkp}$ and (\ref{cond2}), the Lagrangian $L(\cdot, \Pb, \Qb)$ is concave in $W$ and its maxima are 
    also stationary points, i.e., they are obtained by setting equal to zero the first variation of $L$ along an arbitrary direction $\delta W\in\Ltwomat$:
   \al{\delta L &(W,\Pb, \Qb;\delta W)\nn\\ 
   &= \tr \int(\delta W+\delta W^*)P     -  \delta WW^*Q -W\delta W^*Q\nn\\ 
&= 2 \Re \left(\tr  \int P\delta W^*-QW\delta W^*\right).
 } 
Clearly, in order to have
 $\delta L (W,\Pb, \Qb;\delta W)=0$ for any $\delta W\in\Ltwomat$, we must set 
 \al{ QW-P\equiv O.}
 Therefore, {the} unconstrained maximizers of the Lagrangian are {of the form}
  \al{\label{optimal_W} W^\circ=Q^+P+X}
where $(\cdot)^+$ stands for the Moore--Penrose pseudoinverse, and $\mathrm{Range}\,X(e^{i\vartheta})\subseteq [\mathrm{Range}\,Q(e^{i\vartheta})]^\bot$ for all $\vartheta\in\Tbb$, or equivalently, $QX\equiv O$ pointwise. Thus,
the dual function is
\al{ J & (\Pb,\Qb):=L (W^\circ,\Pb, \Qb)\nn\\ 
 &=\tr   \int Q^{+}P^2+\tr\sum_{k=0}^n Q_k^T R_k-2\tr\sum_{\substack{k=1}}^n P_k^T C_k\label{dual_func_original} } 
and  the dual problem is
\begin{subequations}\label{pb_dual}
	\al{&\underset{\Pb,\Qb}{\min}\, J(\Pb,\Qb)\nn\\
		& \hbox{ s.t. }{\Pb\in\Bfko}\nn\\
		& \hspace{.6cm} {\Qb\in\overline{\Bfkp} }\label{dual_constraint_poly}\\
		& \hspace{.6cm} \mathrm{Range}\, P(e^{i\vartheta})\subseteq \mathrm{Range}\, Q(e^{i\vartheta}), \; \forall \, \vartheta\in\Ts. \label{dual_constraint_range}
	}
\end{subequations}
Notice that we have used the stronger constraint \eqref{dual_constraint_range} in place of \eqref{cond2}.  However, it is not obvious that the dual problem admits a solution. Next, we consider a special situation. Let 
\al{\label{Bop} \Bfkop:=\{\Pb\in\Bfko : P>0 \hbox{ on } \Ts\}.}
Let $(\Pb,\Qb)$ be a point in the feasible set of the dual problem (\ref{pb_dual}). According to \eqref{cond4}, if $\Pb\in\Bfkop$, then $\Qb\in\Bfkp$. In this case,
\al{W^\circ=Q^{-1}P\label{optimal_W_special}}
 and the dual function evaluated for $(\Pb,\Qb)\in\Bfkop\times \Bfkp$ becomes
\al{\label{J_special}J  (\Pb,\Qb)=\tr   \int Q^{-1}P^2+\tr\sum_{k=0}^n Q_k^T R_k-2\tr\sum_{\substack{k=1}}^n P_k^T C_k.}
  
\begin{proposition}\label{prop_interior_solution} 
	If the dual problem \eqref{pb_dual} admits a solution $(\Pb,\Qb)\in\mathfrak B_{+}\times \mathfrak B_{o,+}$, 
then $\Phi=Q^{-1}P^2Q^{-1}$ solves the primal problem \eqref{pb_primal}.
\end{proposition}

\begin{proof}  
	The function $J$ in \eqref{dual_func_original} is convex because it is a dual function. Since $(\Pb,\Qb)$ belongs to the interior of the feasible set, it follows that $(\Pb,\Qb)$ must be a stationary point for $J$. Finally, it is not difficult to see that these stationary-point conditions coincide with the moment constraints in (\ref{pb_primal}) with $W=Q^{-1}P$, i.e., the duality gap between the primal and dual problems is equal to zero.
\end{proof}

 \begin{remark}
	If the polynomial matrix $Q$ has only isolated singularity, i.e. if $\det Q$ is not identically zero on $\Tbb$, then the optimal spectral factor in \eqref{optimal_W_special} is still valid almost everywhere. Consequently, the formula for the dual function \eqref{J_special} makes sense provided that the zeros of $\det Q$ (which form a finite set) are excluded from the domain of integration. In this case, however, the dual function should be interpreted as an \emph{extended real-valued} function, and its values at boundary points should be defined with care. This point will be made clear in the next section.
\end{remark}

\section{Regularized dual problem}\label{sec:reg_dual_prob}

As pointed out in the previous section, when $\Pb\in\Bfkop$, any feasible point $(\Pb,\Qb)$ of the dual problem belongs to 
$\Bfkop\times \Bfkp$ where we have $\mathrm{Range}\,P(e^{i\vartheta})=\mathrm{Range}\,Q(e^{i\vartheta}) = \Cbb^m$ for all $\vartheta\in\Tbb$,  and  the dual function takes the simpler form (\ref{J_special}). The idea is to consider a regularized version of the dual problem that explicitly enforces $\Pb$ to belong to $\Bfkop$:
\al{\label{pb_dual_reg}&\underset{\Pb,\Qb}{\min}\, \tilde J_\tau(\Pb,\Qb)\nn\\
& \hbox{ s.t. } \Pb\in\Bfkop,\; \Qb\in {\Bfkp},
}
where
\al{\label{reg_dual_func} & \tilde J_\tau (\Pb,\Qb):=J(\Pb,\Qb)+\tau \tr\int P^{-1}\nn\\
&=\tr\!  \int Q^{-1}P^2+\!\tr\sum_{k=0}^n Q_k^T R_k-\!2\tr\sum_{\substack{k=1}}^n P_k^T C_k+\!\tau \tr \!\int P^{-1} }
and $\tau>0$ is a regularization parameter.

As pointed out in the proof of Proposition~\ref{prop_interior_solution}, the dual function $J$ is convex. Based on this fact, it is not difficult to show the convexity of the regularized dual function $\tilde{J}_\tau$. However, a direct analysis with $\tilde{J}_\tau$ can yield strict convexity, which is a stronger result. We will need the next lemma.
	
	\begin{lemma}\label{lem_g_strict_conv}
	Fix a real number $\tau>0$. For two (Hermitian) positive definite matrices $A$ and $B$, define a real-valued function
	\begin{equation}\label{func_g}
		g(A,B) := \tr(B^{-1}A^2) +\tau \tr(A^{-1}). 
	\end{equation}
	Then $g$ is strictly convex in $(A,B)$.
\end{lemma}

\begin{proof}
	Let us compute the first- and second-order differentials of $g$:
	\begin{equation}\label{first_diff_g}
		\begin{aligned}
			\delta g(A, B; \delta A, \delta B) & = \tr\left[-B^{-1}\delta B B^{-1}A^2 \right.\\
			& \quad \left.+ B^{-1}(\delta A A+A\delta A)-\tau A^{-1}\delta A A^{-1}\right],
		\end{aligned}
	\end{equation}
	\begin{equation}\label{sec_diff_g}
		\begin{aligned}
			\delta^2 g(A, B; \delta A_1, \delta B_1, \delta A_2, \delta B_2) = \tr\left[ B^{-1}(Y+Y^*) \right. \\
			\left. +\tau(Z+Z^*)\right], 
		\end{aligned}
	\end{equation}
where $Y:=(\delta B_1B^{-1}A-\delta A_1)(AB^{-1}\delta B_2-\delta A_2)$ and $Z:=A^{-1}\delta A_2 A^{-1} \delta A_1 A^{-1}$.
	By letting $(\delta A_1, \delta B_1)=(\delta A_2, \delta B_2)=(\delta A, \delta B)$, we obtain the second variation
	\begin{equation}
		\begin{aligned}
			& \delta^2 g(A, B; \delta A, \delta B) \\
			= & 2\tr\left[ (AB^{-1}\delta B-\delta A)B^{-1}(\delta BB^{-1}A-\delta A) \right. \\
			& \left.+\tau A^{-1}\delta A A^{-1} \delta A A^{-1} \right]\geq 0,
		\end{aligned}
	\end{equation}
where the last inequality holds because both $A^{-1}>0$ and $B^{-1}>0$. Now if we impose $\delta^2 g(A, B; \delta A, \delta B)=0$, then the trace of both terms in the bracket above must vanish, implying $\delta AA^{-1}={0}$ and $\delta BB^{-1}A-\delta A={0}$ which further yield $\delta A=\delta B={0}$. Hence, the second variation of $g$ is positive definite which leads to strict convexity.	
\end{proof}

\begin{proposition}\label{prop_strict_convex} 
	The regularized dual function $\tilde J_\tau$ is strictly convex in $\Bfkop\times \Bfkp$. 
\end{proposition}

 \begin{proof} 
 	The claim can be proved by showing the strict convexity of the function
 \al{f_\tau(\Pb,\Qb) :=\tr   \int Q^{-1}P^2+\tau \tr\int P^{-1} } 
 since $\tilde J_\tau$ and $f_\tau$ only differs from linear terms in $(\Pb,\Qb)$. 
 Using the function $g$ defined in \eqref{func_g}, we can rewrite	 $f_\tau$ as $\int g(P(e^{i\vartheta}), Q(e^{i\vartheta}))$. Furthermore, for any $t\in(0, 1)$ and any two points $(\Pb_1, \Qb_1)$ and $(\Pb_2, \Qb_2)$ in $\Bfkop\times\Bfkp$, Lemma~\ref{lem_g_strict_conv} implies that
 	\begin{equation}
 		\begin{aligned}
 			& f_\tau\left(t(\Pb_1, \Qb_1) + (1-t)(\Pb_2, \Qb_2)\right) \\
 			= & \int g\left( t(P_1(e^{i\vartheta}), Q_1(e^{i\vartheta})) + (1-t) (P_2(e^{i\vartheta}), Q_2(e^{i\vartheta}))\right) \\
 			< & \, t \int g\left(P_1(e^{i\vartheta}), Q_1(e^{i\vartheta})\right) + (1-t) \int g\left(P_2(e^{i\vartheta}), Q_2(e^{i\vartheta})\right) \\
 			= &\, t f_\tau\left(\Pb_1, \Qb_1\right) + (1-t)f_\tau\left(\Pb_2, \Qb_2\right),
 		\end{aligned}
 	\end{equation}
 	which is by definition the strong convexity of $f_\tau$.
 	The strict inequality holds in the above formula because of the continuity of the integrands.
  \end{proof}

\textbf{Feasibility assumption:} There exists a spectral density $\bar \Phi$ which is positive definite on an open interval in $\Ts$ and satisfies the moment constraints
\al{\label{feas_cond}\int e^{i\vartheta k} \bar \Phi=R_k,\quad k= 0, \ldots, n.}

Notice that under the feasibility assumption, we have 
\al{\label{feas_cond_conseq}\tr\sum_{k=0}^n Q_k^T R_k=\tr\int \bar \Phi Q\geq 0.}
In what follows, we always suppose that the feasibility assumption holds. 
In practice, the feasibility assumption can be guaranteed by suitably estimating the covariances using e.g., the periodogram; see \cite{ZFKZ2019M2,FALCONI2023110672}. This point will be made precise in the next section.

Next, we shall characterize the behavior of $\tilde{J}_\tau$ near 
	the boundary of the feasible set.
	Before to proceed, let us define the norm of $\Pb$ and $\Qb$ as
	\al{\|\Pb\|^2 & := \frac{1}{2} \sum_{k=1}^{n} \tr (P_kP_k^T) = \int \|P(e^{i\vartheta})-I\|_{\F}^2, \label{def_P_norm} \\
		\|\Qb\|^2 & := \tr (Q_0Q_0^T) +\frac{1}{2}\sum_{k=1}^{n} \tr (Q_kQ_k^T) = \int \|Q(e^{i\vartheta})\|_{\F}^2, \label{def_Q_norm} }
	and $\|(\Pb,\Qb)\|^2:=\|\Pb\|^2+\|\Qb\|^2$. Here the factor $1/2$ comes from the definition of the polynomial matrices in \eqref{def_poly_P_Q}, and $\|\cdot\|_{\F}$ denotes the Frobenius norm of a matrix.
	The following lemma will be useful in the subsequent development.
	\begin{lemma}\label{lem_P_bounded}
		The set $\Bfkop$ in \eqref{Bop} is bounded, i.e., there exists a constant $C>0$ such that any $\Pb\in\Bfkop$ satisfies $\|\Pb\|\leq C$.
	\end{lemma}
	
	\begin{proof}
		Let us take $\Pb\in\Bfkop$.
		The following computation is straightforward:
		\begin{equation}\label{form51}
			\begin{aligned}
				\|P_k\|_{\F}^2 & = \tr(P_kP_k^T) = \tr\left[\left(2{\int}  e^{i\vartheta k}P\right) \left(2{\int} e^{-i\varphi k}P\right)\right] \\
				& = 4{\int} {\int}e^{i(\vartheta-\varphi)k} \tr\left[P(e^{i\vartheta})P(e^{i\varphi})\right], \\
			\end{aligned}
		\end{equation}
		where {the two integrals are taken with respect to different integration variables, $\vartheta\in\Tbb$ and $\varphi\in\Tbb$. The last trace term in \eqref{form51}} is nonnegative for any $\vartheta$ and $\varphi$. Taking the modulus (of a complex number) on both sides of the equation, we have
		\begin{equation}
			\begin{aligned}
				\|P_k\|_{\F}^2 & \leq 4 {\int \int} \tr\left[P(e^{i\vartheta})P(e^{i\varphi})\right] \\
				& = 4 {\int \int} \sum_{i,\, j=1}^m p_{ij}(e^{i\vartheta}) p_{ij}^*(e^{i\varphi}) \\
				& = 4 \sum_{i,\, j=1}^m {\int} p_{ij}(e^{i\vartheta}) {\int}   p_{ij}^*(e^{i\varphi}) =4m,
			\end{aligned}
		\end{equation}
		where the last equality follows from the fact that
		\begin{equation*}
			\int p_{ij} = \left\{\begin{array}{ll}
				1 & \text{if}\ i=j \\
				0 & \text{if}\ i\neq j
			\end{array}\right.
		\end{equation*}
		is the constant term of the scalar polynomial $p_{ij}$, the $(i, j)$ entry of the polynomial matrix $P$ whose constant term is $I$. In view of \eqref{def_P_norm}, we have $\|\Pb\|\leq \sqrt{2mn}=:C$.
	\end{proof}

\begin{proposition} \label{first_prop}
Consider a sequence $\{(\Pb^l,\Qb^l)\}_{l\geq 1}\subset\Bfkop\times \Bfkp$
 that converges to a point
 $(\bar\Pb,\bar\Qb)\in \partial(\Bfkop\times \Bfkp)$ as $l$ diverges. Then, 
 \al{\lim_{l\rightarrow \infty} \tilde J_\tau(\Pb^l,\Qb^l)=\infty,}
 i.e., $\{(\Pb^l,\Qb^l)\}_{l\geq 1}$ is not an infimizing sequence.   
\end{proposition} 
\begin{proof} Since $\Pb^l\in \Bfkop$, we know that $\Pb^l$ is bounded by Lemma~\ref{lem_P_bounded}, and this further implies that the inner product $\tr\sum_{k=1}^n (P_k^l)^T C_k$ is also bounded. Let us first consider the case in which $(\bar\Pb,\bar\Qb)\in \partial\Bfkop\times \overline\Bfkp$.
In view of (\ref{feas_cond_conseq}) we have that 
\al{\label{ineq_1}\tilde J_\tau(\Pb^l,\Qb^l)\geq   -2\tr\sum_{k=1}^n (P_k^l)^T C_k+\tau \int\tr[( P^l)^{-1}]. }
The term $\tr[( P^l)^{-1}]$ is a nonnegative rational function and it has at least one pole tending to the unit circle as $l\rightarrow\infty$. Accordingly, the second term in (\ref{ineq_1}) tends to infinity as $l\rightarrow \infty$. Moreover, the latter dominates the first term in \eqref{ineq_1} and thus we conclude that $\tilde J_\tau$ diverges when $\Pb^l$ tends to $\bar \Pb \in \partial\Bfkop$. 

It remains to show that $\tilde J_\tau$ also diverges when $(\bar \Pb, \bar\Qb)\in\Bfkop\times \partial \Bfkp$. Let $\varepsilon_l\geq 0$ denote the minimum over $\Ts$ of the minimum eigenvalue of the polynomial matrix $P^l(e^{i\vartheta})$. Since the latter converges to $\bar P\in\Bfkop$, there exist two constants $\epsilon>0$  and $\bar l\in\mathbb N$ such that $\varepsilon_l\geq \varepsilon$  for $l\geq \bar l$. Therefore, 
\al{\tr\int  (Q^l)^{-1} (P^l)^2\geq \varepsilon^{2} \tr\int (Q^l)^{-1}\nn,}
and thus 
\al{\label{ineq_2}\tilde J_\tau(\Pb^l,\Qb^l)\geq   \varepsilon^{2} \int \tr[(Q^l)^{-1}] -2\tr\sum_{\substack{k=1}}^n (P_k^l)^T C_k,}
where the first term on the right-hand side dominates the second one because it diverges since 
 $\tr[(Q^l)^{-1}]$ is a nonnegative rational function having at least one pole tending to the unit circle as $l\rightarrow \infty$. 
\end{proof}

The next  proposition shows the behavior of $\tilde{J}_\tau$ when the norm of its argument tends to infinity.

\begin{proposition}\label{second_prop}
Consider a sequence $\{(\Pb^l,\Qb^l)\}_{l\geq 1}\subset \Bfkop\times \Bfkp$
 such that $\|(\Pb^l,\Qb^l)\|\to \infty$ as $l\rightarrow \infty$. Then, 
 \al{\lim_{l\rightarrow \infty} \tilde J_\tau(\Pb^l,\Qb^l)=\infty.}
\end{proposition}
\begin{proof} 
According to Lemma~\ref{lem_P_bounded}, $\Pb^l$ is bounded in the feasible set and thus $\|(\Pb^l,\Qb^l)\|\rightarrow \infty$  implies that $\|\Qb^l\|\rightarrow \infty$. By \eqref{feas_cond_conseq}, we have
\al{\label{inequal_inner_prod}
	\tilde J_\tau(\Pb^l,\Qb^l)&\geq \tr\sum_{k=0}^n (Q_k^l)^T R_k-2\tr\sum_{\substack{k=1}}^n (P_k^l)^T C_k \nn\\
&= \tr\int \bar \Phi Q^l -2\tr\sum_{\substack{k=1}}^n (P_k^l)^T C_k}
where the second term remains bounded as $l\rightarrow \infty$. Let $\tilde\Qb^l:= \Qb^l/\|\Qb^l\|$ and take a subsequence $\tilde\Qb^{l_s}$, with
$s\in\mathbb N$, such that $\lim_{s\to\infty}\tr\int \bar \Phi \tilde Q^{l_s}=\eta :=\liminf_{l\to\infty}\tr\int \bar \Phi \tilde Q^{l}\geq 0 $.
 Since such a subsequence belongs to the surface of the (compact) unit ball, then there exists a convergent subsubsequence $\tilde\Qb^{l_t}$, with $t\in\mathbb N$ and let 
\al{\tilde \Qb=\lim_{t\rightarrow \infty} \tilde \Qb^{l_t}\in\overline{\Bfkp}.} 
Then, $\eta=\tr\int \bar \Phi \tilde Q$ and we claim that $\eta >0$. Indeed, assume by contradiction that $\eta=0$. Since $\tr(\bar\Phi \tilde Q)\geq 0$ a.e. on $\Ts$, $\eta=0$ implies that $\tr(\bar\Phi \tilde Q )= 0$ a.e. on $\Ts$. By the feasibility assumption, $\tilde Q$ must vanish in some open interval in $\Ts$ and thus $\tilde \Qb\equiv 0$. This is a contradiction because $\|\tilde \Qb\|=1$. We conclude that $\eta>0$. 
Since by definition $\eta=\liminf_{l\to\infty}\tr\int \bar \Phi \tilde Q^{l}$, there exists a positive integer $\bar l$ such that $\tr\int \bar \Phi \tilde Q^{l} >\eta/2$ for all $l\geq \bar l$.
Consequently, 
\al{
	\tr\int \bar \Phi Q^l & = 
	 \|\Qb^{l}\| \tr\int \bar \Phi \tilde Q^{l} \nn\\ 
	 & >\frac{\eta}{2} \|\Qb^{l}\| \to \infty \ \text{as}\ l\to\infty,  }
	 where the inequality holds for all $l\geq \bar l$. 
	In view of \eqref{inequal_inner_prod}, the above formula implies $\tilde J_\tau(\Pb^{l},\Qb^{l})\rightarrow \infty$, i.e., $\{(\Pb^l,\Qb^l)\}_{l\geq 1}$ is not an infimizing sequence.

 \end{proof}

\begin{theorem}\label{thm_main}
The regularized dual problem \eqref{pb_dual_reg} admits a unique solution, say $(\Pb,\Qb)$. Moreover, the spectral factor $W^\circ=Q^{-1}P$ satisfies the moment  conditions:
\al{\label{moment_constr2}&\int e^{i\vartheta k}  W^\circ(W^\circ)^*=R_k,\quad k=0, \ldots, n\nn \\
 & \frac{1}{2} \int e^{i\vartheta k}   \left[ W^\circ+(W^\circ)^* \right]=C_k+E_k,\quad k=1, \ldots, n}
 with the error of cepstral matching
 \al{E_k:=\frac{\tau}{2}\int e^{i\vartheta k}  P^{-2}.\nn}
\end{theorem}
\begin{proof} 
	Consider the sublevel set of the regularized dual function
	\begin{equation}
		\tilde{J}_\tau^{-1}(-\infty, \beta]:=\{(\Pb, \Qb) \in \Bfkop\times \Bfkp : \tilde{J}_\tau (\Pb, \Qb)\leq \beta\}
	\end{equation}
	where $\beta\in\Rbb$ is sufficiently large such that the set above is nonempty. For example, one can take $\beta=\tilde{J}_\tau (\Pb_0, \Qb_0)$ for an arbitrarily chosen $(\Pb_0, \Qb_0) \in \Bfkop\times \Bfkp$. The aim is to show that the sublevel set is \emph{compact}, i.e., closed and bounded because $(\Pb, \Qb)$ resides in a finite-dimensional space. For the closedness, suppose that there is a sequence $\{(\Pb^l, \Qb^l)\}_{l\geq 1}\subset \tilde{J}_\tau^{-1}(-\infty, \beta]$ that converges to $(\bar \Pb, \bar \Qb)\in\overline \Bfkop\times \overline\Bfkp$. We need to show that $(\bar \Pb, \bar \Qb)$ are in fact in the open set $\Bfkop\times \Bfkp$ so that $(\bar \Pb, \bar \Qb)$ belongs to the sublevel set by continuity. To this end, suppose that $(\bar \Pb, \bar \Qb)\in \partial (\Bfkop\times \Bfkp)$. Then according to Proposition~\ref{first_prop}, the sequence of function values $\tilde{J}_\tau(\Pb^l,\Qb^l)$ will eventually become larger than $\beta$, which means that $(\Pb^l,\Qb^l)\notin \tilde{J}_\tau^{-1}(-\infty, \beta]$ for all sufficiently large $l$, a contradiction. Hence the limit point $(\bar \Pb, \bar \Qb)$ must be an interior point and also a member of the sublevel set. The boundedness of $\tilde{J}_\tau^{-1}(-\infty, \beta]$ can be proved in a similar way with reference to Proposition~\ref{second_prop}.
	 Accordingly, the regularized dual problem admits a point of minimum on the compact sublevel set by Weierstrass' theorem. The uniqueness follows from the fact that $\tilde J_\tau$ is strictly convex in $\Bfkop\times \Bfkp$ (Proposition~\ref{prop_strict_convex}). Finally, such a minimizer must be a stationary point of $\tilde J_\tau$. The stationarity conditions are obtained by imposing 
\al{ & \delta \tilde J_\tau(\Pb,\Qb,\delta\Pb,\delta\Qb) \nn \\
	= & \, 2\Re\,\tr \left\{ \frac{1}{2} \sum_{k=0}^{n} \delta Q_k^T \left(R_k -\int e^{i\vartheta k} Q^{-1}P^2Q^{-1}\right) \right.\nn \\
	& \left. + \sum_{k=1}^n\delta P_k^T\left[\frac{1}{2}\int e^{i\vartheta k} (PQ^{-1} +Q^{-1}P -\tau P^{-2}) -C_k \right]\right\} \\
	= & \, 0 , \;\forall \, \delta\Pb\in\Bfko,\; \delta\Qb\in\Bfk, \nn}
which clearly implies
 the moment constraints in (\ref{moment_constr2}) with $W^\circ=Q^{-1}P$.
\end{proof}

\section{Spectral estimation}\label{sec:est}

Let $y_1, \ldots, y_N$ be a finite-length realization obtained from a process with spectral density $\Phi$. We can compute a coarse estimate of $\Phi$ from the data by means of the $f$-periodogram \cite{FALCONI2023110672}:
\al{\label{fper}\hat \Phi_N (e^{i\vartheta}) = \sum_{k=-f(N)}^{f(N)} e^{-i\vartheta k} \hat R_k}
where 
\al{\label{cov_est}\hat R_k=\frac{1}{N}\sum_{t=1}^{N-k}y_{t+k} y_t^T}
is the \emph{standard biased covariance estimator}.  Moreover, we assume that $\hat \Phi_N$ is positive definite a.e. on $\Tbb$. {Let $\hat W_N$ be a canonical left  spectral factor\footnote{By ``canonical'', we mean that {$\hat W_N$ is univocally determined by $\hat \Phi_N$.} 
}  of the $f$-periodogram, and thus  $\hat \Phi_N=\hat W_N\hat W_N^*$. We can take $\hat W_N$ as the causal and minimum phase spectral factor which is} of the form
\al{\label{W_hat_N}\hat W_N (e^{i\vartheta})=\sum_{k=0}^{f(N)} e^{-i\vartheta k}\hat S_k.} {Another possible choice is to take $\hat W_N$ as the pointwise Cholesky factor, i.e., $\hat W_N$ is a continuous matrix  function such that $\hat W_N(e^{i\vartheta})$ is lower triangular and $\hat W_N(e^{i\vartheta}) \hat W_N(e^{i\vartheta})^*=\hat \Phi_N(e^{i\vartheta})$ for all $\vartheta$.} Notice that, the spectral factor is not unique. More precisely, all the left spectral factors of $\hat \Phi_N$ are obtained from $\hat W_N$ as
\al{ \hat W=\hat W_NU}
where $U$ is an inner matrix function, i.e. $U(e^{i\vartheta})U(e^{i\vartheta})^*=I$ for all $\vartheta\in\Ts$.  For the moment, we assume that $U$ is fixed.  Then, an estimator of $\Phi$ is obtained by solving the regularized dual problem (\ref{pb_dual_reg}) where $R_k$ is replaced by $\hat R_k$ and $C_k$ by
\al{\label{chat}\hat C_k=\frac{1}{2}\int e^{i\vartheta k} (\hat W_N U+U^*\hat W^*_N).}
It is worth noting that the feasibility condition \eqref{feas_cond} is satisfied by taking $\bar\Phi= \hat \Phi_N$. By (\ref{fper}) and (\ref{chat}), the regularized dual function in \eqref{reg_dual_func} can be rewritten as
\al{&\tilde J_\tau(\Pb,\Qb, U) = \nn\\ 
&\,\tr   \int \left[ Q^{-1}P^2+ Q \hat\Phi_N-(P-I)(\hat W_N U+U^*\hat W^*_N)+\tau P^{-1}\right]\label{J_tilde_P_Q_U}}
where we have explicitly included the inner function $U$ as an {additional} argument.
Then, the minimization of $\tilde J_\tau$ with respect to $(\Pb,\Qb)$ is equivalent to minimizing the function
\al{&\check J_\tau(\Pb,\Qb)=\tilde J_\tau(\Pb,\Qb)- \tr\int \left(\hat W_N U+U^*\hat W^*_N \right)\nn\\ 
&=\tr   \int \left[ Q^{-1}P^2+ Q \hat\Phi_N-P(\hat W_N U+U^*\hat W^*_N)+\tau  P^{-1} \right] \nn}
as the two functions differ from a term not depending on $(\Pb,\Qb)$. Moreover, notice that
\al{\label{jcheck}&\check J_\tau(\Pb,\Qb)\nn\\
= &\,\tr \! \int \left[ P^2Q^{-1}+ Q \hat\Phi_N-Q\hat W_N UPQ^{-1}-PU^*\hat W^*_N+\tau  P^{-1} \right]\nn\\
= &\,\tr  \! \int\! \left[ Q(Q^{-1}P^2Q^{-1}+ \hat\Phi_N- \hat W_N UPQ^{-1}-Q^{-1}PU^*\hat W^*_N) \right.\nn\\
&\hspace{.5cm} \left.+\tau P^{-1} \right] = \int \|Q^{-1}P- \hat W_N U\|_Q^2+ \tau\trace\int P^{-1}, 
}
where $\|X\|_Q:=\tr(X^*QX)$ with matrices $X$ and $Q$ of compatible size and $Q>0$.
In plain words, the dual problem finds the spectral factor $W=Q^{-1}P$ which is the closest to $\hat W_N U$ according to a weighted norm. As we already noticed, such solution depends on the particular spectral factor that we have chosen, i.e., it depends on the inner function $U$. Moreover, $\check J_\tau(\Pb,\Qb)\geq 0$ and, neglecting the effect of the regularization term, $\check J_\tau(\Pb,\Qb)\approx 0$ only when $Q^{-1}P\approx \hat W_N U$. 
Note that, if we take $U\equiv I$, then $\hat W_N U$ is a causal spectral factor, while $Q^{-1}P$ is not. Accordingly, the idea is to learn jointly $(\Pb,\Qb)$ and $U$ through an alternating minimization scheme. 
Then, at Stage $\ell\in\Nbb_+$ of the procedure we perform the following steps:\\
$\bullet$ Spectral factor optimization 
\al{\label{step1}&(\Pb^{(\ell)},\Qb^{(\ell)})=\underset{\Pb,\Qb}{\mathrm{argmin}}\, \check J_\tau(\Pb,\Qb,U^{(\ell-1)})\nn\\
&\hspace{2.5cm} \hbox{ s.t. } \Pb\in\Bfkop,\; \Qb\in {\Bfkp}
}
$\bullet$ Inner function optimization
 \al{\label{step2}& U^{(\ell)}=\underset{U}{\mathrm{argmin}}\, \check J_\tau(\Pb^{(\ell)},\Qb^{(\ell)},U)\nn\\
&\hspace{1.5cm} \hbox{ s.t. } UU^*=I \quad \forall \, \vartheta\in\Ts.
}
The alternating minimization procedure {can be} initialized with $U^{(0)}\equiv I$.
Clearly, the problem (\ref{step1}) admits a unique global minimum. If also (\ref{step2}) admits a global minimum, and if $(\Pb^{(\ell)},\Qb^{(\ell)},U^{(\ell)})$ converges,
then such a limit is a coordinatewise point of minimum for $\check J_\tau(\Pb,\Qb,U)$. A point of  minimum of \eqref{step2} can be characterized as follows. 
	\begin{proposition}\label{prop_update_U_inn}
		A solution to the minimization problem \eqref{step2} is 
		\al{\label{optU}U(e^{i\vartheta})=\hat V(e^{i\vartheta})V(e^{i\vartheta})^*}
		where 
		\al{P^{(\ell)}(e^{i\vartheta})\hat W_N(e^{i\vartheta})=V(e^{i\vartheta}) D(e^{i\vartheta}) \hat V(e^{i\vartheta})^*\label{svd_PW_hat}}
		is a pointwise singular value decomposition of $P^{(\ell)}\hat W_N$ with $VV^*=\hat V\hat V^*=I$ and $D >0$  diagonal. 
	\end{proposition}
	\begin{proof}
		By \eqref{jcheck}, the optimization problem (\ref{step2}) is equivalent to minimizing 
		\al{-\tr   \int \hat W_N UP^{(\ell)}+P^{(\ell)}U^*\hat W^*_N=-2  \Re\,\tr \int P^{(\ell)}\hat W_N U,\nn}
		which is the same as maximizing $\Re\,\tr \int P^{(\ell)}\hat W_N U$		
		under the constraint $UU^*=I$. 
		In view of \eqref{svd_PW_hat}, we have the following chain of inequalities:
		\begin{equation}
			\begin{aligned}
				\Re\,\tr \int P^{(\ell)}\hat W_N U & \leq \left| \tr \int D \tilde{U} \right|\\
				& \leq \int \left| \tr (D \tilde{U}) \right| \leq \int \tr D 
			\end{aligned}
		\end{equation}
		where the first inequality comes from the fact $\Re\, z\leq |z|$ for any $z\in\Cbb$ and $\tilde{U}:=\hat V^* U V$ is pointwise unitary.
        Clearly, all the inequalities above become equalities if one takes $\tilde{U}\equiv I$.
		Therefore, a minimizer of \eqref{step2}
        is (\ref{optU}) and it is an all-pass function because it is a product of two such functions $\hat{V}$ and $V$. 
	\end{proof}

	\begin{remark}\label{rem_update_C}
		In practice, upon the completion of Stage $\ell$, we can compute the values of the optimal solution $U^{(\ell)}$ to \eqref{step2} over a grid in $[0,2\pi)$ and then we use these values to update the $W$-cepstral coefficients through
		\al{\label{hat_C_k_updated} \hat C_k^{(\ell+1)}=\frac{1}{2}\int e^{i\vartheta k} (\hat W_N U^{(\ell)}+(U^{(\ell)})^*\hat W_N^*),\quad k=1, \dots, n}
		where the integral is approximated with a (finite) Riemann sum. Then in Stage $\ell+1$, we can solve the minimization problem \eqref{step1} with the updated $W$-cepstral coefficients $\{\hat C_k^{(\ell+1)}\}_{k=1}^n$.
\end{remark}

Next, we analyze the consistency of the proposed estimator. 


\textbf{Data generation assumption:} The data $y_1, \ldots, y_N$ have been generated by a process having spectral density 
\al{ \Phi_{\text{true}}= Q_{\text{true}}^{-1}P_{\text{true}}^2Q_{\text{true}}^{-1}}
with $\Pb_{\text{true}}\in\Bfkop$ and $\Qb_{\text{true}}\in \Bfkp$. \smallskip\smallskip

Let ${\cal A}_{m,+}$
denote the set of $\Cbb^{m\times m}$-valued functions of $z\in\Cbb$ which are analytic in an annulus containing the unit circle and
are positive definite for all $z=e^{i\vartheta}$ with $\vartheta\in\Tbb$.
Consider the continuous map 
 for constructing a left canonical spectral factor of an element  in ${\cal A}_{m,+}$:  
\al{{\cal I} : \;{\cal A}_{m,+} &\rightarrow \Ls^2_{m\times m}(\Ts)\nn \\   \Phi &\mapsto W \nn}
such that $\Phi=WW^*$ and
\al{& W(z)=\sum_{k=0}^\infty W_k z^{-k}\nn}
{is analytic outside a disk with radius smaller than one.}
 Since $ \Phi_{\text{true}}$ is positive definite in $\Tbb$, the true process can be generated by a bounded input-bounded output (BIBO) stable system fed by white noise. Choosing $f(N)$ such that $f(N)\rightarrow \infty$ and  $f(N)/\sqrt{N} \rightarrow 0$ as $N\rightarrow\infty$, by \cite[Theorem 1]{FALCONI2023110672} we have that
\al{ \hat \Phi_N \xrightarrow{\text{m.s.}} \Phi_{\text{true}} \quad \text{as}\quad N\to\infty,
}
where m.s.~indicate mean square convergence.
 Therefore, for $N$ sufficiently large $\hat W_N={\cal I}(\hat \Phi_N)$ is well defined and by the 
 continuous mapping theorem 
 we have that 
\al{\hat W_N \xrightarrow{\text{m.s.}} W_{\text{true}} \quad \text{as}\quad N\to\infty\nn}
with $W_{\text{true}}={\cal I}(\Phi_{\text{true}})$.
Let $\check J_{0,N}(\Pb,\Qb,U)$ 
denote the function in 
\eqref{jcheck} with $\tau=0$. Then, {taking the limit  $N\to\infty$} we obtain
\al{ \check J_{0,\infty}(\Pb,\Qb,U)=  \int \|Q^{-1}P-W_{\text{true}} U\|_Q^2\geq 0}
and it is not difficult to see that
\al{\check J_{0,\infty}(\Pb_{\text{true}},\Qb_{\text{true}},U^\star)=0}
where $\Pb_{\text{true}},\Qb_{\text{true}}$ correspond to $P_{\text{true}},Q_{\text{true}}$, respectively, and $U^\star = W_{\text{true}}^{-1} Q_{\text{true}}^{-1}P_{\text{true}} = W_{\text{true}}^{-1} W^\circ_{\text{true}}$ is all pass. 
Notice that passing to the limit as $N\to\infty$ in the above equalities involves an interchange of the limit with the integral which can be justified by the dominated convergence theorem. In fact, one can argue along the lines of the proof of Theorem~\ref{thm_main} to show that the eventual optimal solution $(\Pb^{(\infty)}, \Qb^{(\infty)})$ must remain in a compact (sublevel) set.
In summary, without regularization a 
global  point of minimum for {$\check J_{\tau, \infty}$} corresponds to the true model.

%
%

\section{Simulations}\label{sec:sims}

In this section, we present some numerical simulations to {assess the performance of the spectral estimator proposed in Section \ref{sec:est}}. 
More precisely, we implement an estimation procedure starting from a finite sample path of a stationary process and test the alternating minimization scheme \eqref{step1} and \eqref{step2}. 
Notice that the numerical solution of the regularized dual problem \eqref{pb_dual_reg} {by means of} Newton's method was presented in the conference paper \cite[Sec.~4]{Zhu-Zorzi-2025} along with many {implementation details}, including the construction of a basis for the vector space containing the dual variables, the computation of the gradient and the Hessian of the regularized dual function, and the initialization of the dual variables. Therefore, these details are omitted here.

\subsection{Parameters setting}\label{subsec:para_set}

%
%

In order to showcase the spectral estimation procedure described in Section~\ref{sec:est}, we consider a {vector-valued} process $\{y_t\}_{t\in \Zbb}$ with $m=3$ scalar components
	which admits  the unilateral ARMA representation of order two:
\begin{equation}\label{ARMA_causal}
		\sum_{k=0}^2 A_ky_{t-k}
		= \sum_{k=0}^{2} B_kw_{t-k},\quad t\in\Zbb
\end{equation}
where
{ 	\begin{equation}\label{param_ARMA}
		\begin{split}
			A_0 & =\bmat 
			2.39       &  0&         0\\
			-0.28&    2.52     &    0\\
			-0.17 &  -0.32   & 0.97
			\emat,\\ 
			A_1  &=\bmat 
			0.04 &  -1.04 &  -0.08\\
			0.44  &  0.01 &   0.06\\
			0.12  & -0.44  & -0.44
			\emat,   \\
			A_2 & =\bmat 
			-1.13 &   0.44 &  -0.19\\
			-2.55  &  1  & -0.49\\
			0.58  & -0.71  &  0.43
			\emat, \\ B_0 &=  \diag(1, 1, 0.95),\\
			B_1 & = \diag(
			-1.6674 ,
			0.9096 ,
			-0.5213)
			,\ \text{and}  \\
			B_2 &= \diag( 0.9025  ,
			0.9025, 
			0.2375)
		\end{split}
\end{equation}}are the matricial parameters, and $\{w_t\}_{t\in\Zbb}$ is a {two-dimensional} white {Gaussian} noise process of unit variance, i.e., $\E w_t w_t^T=I$. The spectral density of the process $y_t$ is 	  $\Phi^\circ(e^{i\vartheta})= \left.A(z)^{-1}B(z)B(z^{-1})^TA(z^{-1})^{-T}\right|_{z=e^{i\vartheta}}$ with $A(z)=\sum_{k=0}^2 A_kz^{-k}$ and $B(z)= \sum_{k=0}^{2} B_kz^{-k}$.
	The peculiarity of this model is that two polynomials in the main diagonal of $B(z)$ have zeros\footnote{These are the \emph{transmission zeros} at which the spectrum or the transfer function of \eqref{ARMA_causal} loses rank.} very close to the unit circle (which are $0.95e^{\pm.5i}$ and $0.95e^{\pm2.07i}$, respectively). Thus, it is difficult to approximate such process using an AR process. By construction the system \eqref{ARMA_causal}  is BIBO stable. Thus a sample path $y_1, \ldots, y_N$ of $y_t$ of length $N=500$  is generated from \eqref{ARMA_causal} via standard filtering {techniques}. The $f$-periodogram in \eqref{fper} can then be evaluated with $f(N)=\sqrt[3]{N}$.  To ensure that $\hat{\Phi}_N$ after the truncation is positive definite on $\Tbb$, a positive multiple of the identity matrix 
	is added to it whenever necessary. 
	
For the alternating minimization scheme \eqref{step1} and \eqref{step2}, we terminate the stage when the distance between two successive iterates $(\Pb^{(\ell+1)},\Qb^{(\ell+1)})$ and $(\Pb^{(\ell)},\Qb^{(\ell)})$, as measured by the norm of their difference, is less than $10^{-3}$.

\subsection{Simulation results}

We compute an estimate of $\Phi^\circ$ from the sample path using the following four estimators:
\begin{itemize}
\item VARMA which denotes the  {vector-valued} ARMA spectral estimator proposed in Section \ref{sec:est} with $U^{(0)}\equiv I$ and $\hat{W}_N$ in \eqref{W_hat_N} is computed using Bauer's method \cite{sayed2001survey};
\item  VAR-ma which denotes the  {vector-valued} ARMA spectral estimator whose MA part is scalar   \cite{lindquist2013onthemultivariate,lindquist2016modeling,10588982}; {see also Remark \ref{RemSCAL}};
\item VAR which denotes the Hellinger spectral estimator with a noninformative prior $\Psi=I$ \cite{FPR-08,RFP-09,RFP-10-wellposedness} such that the spectrum corresponds to a  {vector-valued} AR model;
\item  VAR$_{\text{ME}}$ which denotes the maximum entropy spectral estimator \cite{burg1975maximum,georgiou2002spectral}, i.e., the spectrum corresponds to a   {vector-valued} AR model;
\end{itemize}
Notice that VAR and VAR$_{\text{ME}}$ use only the  covariances lags $\{\hat R_k\}_{k=0}^n$, while in addition, VARMA  uses the matricial $W$-cepstral coefficients $\{\hat C_k\}_{k=1}^n$ (see Definition~\ref{def:W_ceps}) and VAR-ma uses the scalar cepstral coefficients $\{\hat c_k\}_{k=1}^n$ (i.e., the scalar logarithmic moments mentioned in Remark \ref{RemSCAL}).  The regularization parameter for the cepstral matching  is set as $\tau = 10^{-3}$ for both VARMA and VAR-ma. For all estimators, the number of retained terms $n$, for both covariance lags and cepstral coefficients, is set equal to 2. 
The $W$-cepstral coefficients used by VARMA  are computed via \eqref{hat_C_k_updated} and the scalar logarithmic moments used by VAR-ma are computed by substituting the true spectrum with the $f$-periodogram in the standard formula.
%

\begin{figure}
	\centering
	
	\includegraphics[width=\linewidth]{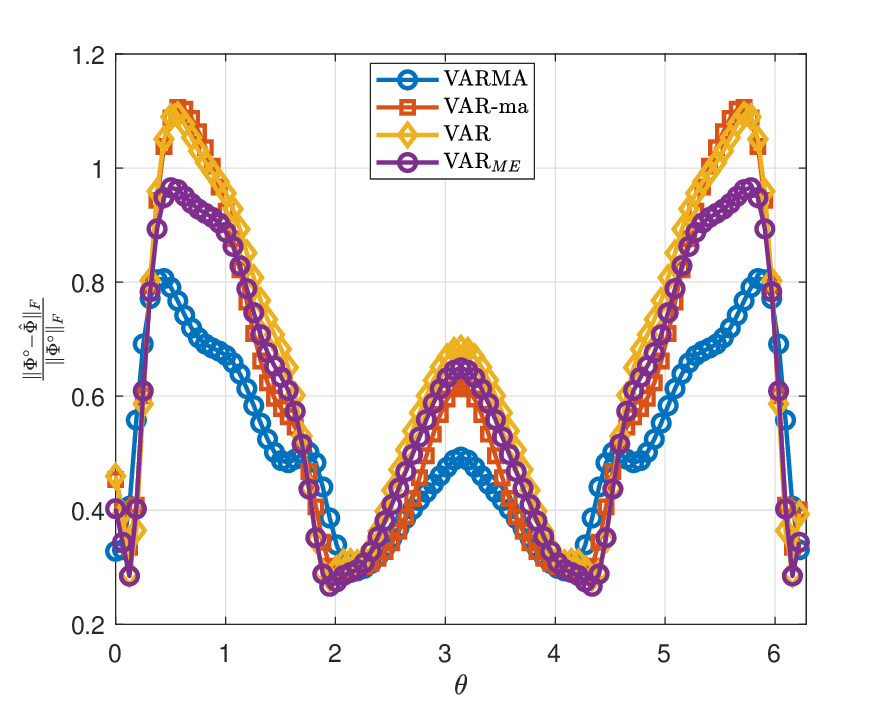}
	\caption{Relative error in the spectrum reconstruction as a function of $\vartheta$ which is averaged over $100$ trials of the Monte Carlo simulation.}		\label{fig:ave_ptw_rel_err_MC}
\end{figure} 
\begin{figure}
	\centering
	
	\includegraphics[width=\linewidth]{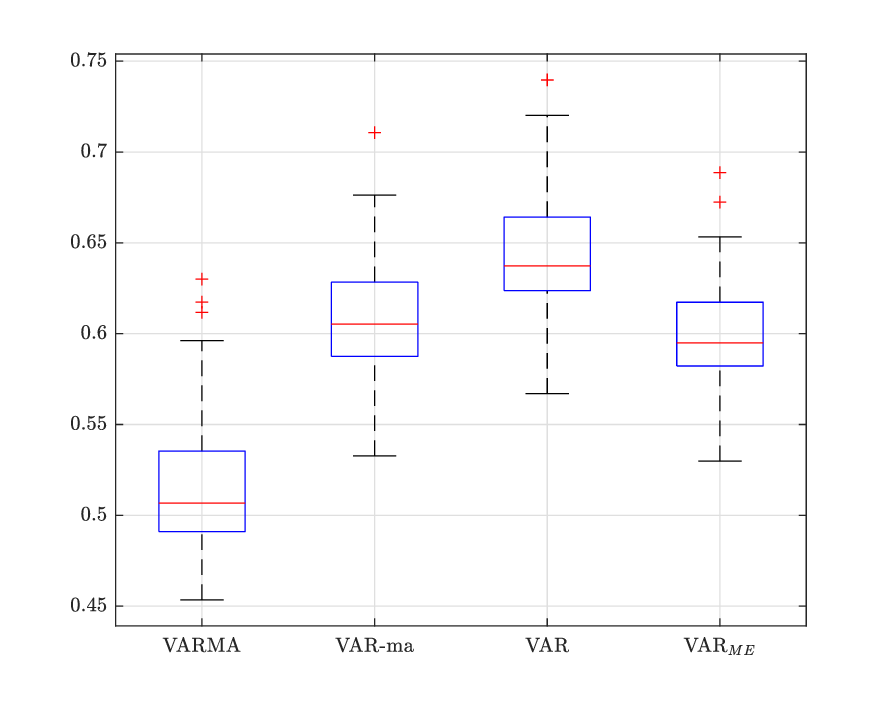}
	\caption{Boxplot of the integrated relative {error for the four  spectral estimators}.}		\label{fig:int_rel_err_MC}
\end{figure}

We perform a Monte Carlo simulation of $100$ trials using the true model \eqref{ARMA_causal} and \eqref{param_ARMA}. In each trial, we generate a sample path of $y_t$ of length $500$ from which we can compute the covariance estimates, the periodogram, and the cepstral coefficients. Then the four spectral estimators listed above are constructed and compared. Two criteria of comparison are adopted. One is the relative error of the reconstructed spectral density $\hat{\Phi}$ which is defined as
	\al{\frac{\|\Phi^\circ(e^{i\vartheta}) -\hat \Phi(e^{i\vartheta})\|_{\F}}{\|\Phi^\circ(e^{i\vartheta})\|_{\F}}\nn}
and is a function of $\vartheta$. 
	The other is called ``integrated relative error'' which is simply 
	\al{\int \frac{\|\Phi^\circ -\hat \Phi\|_{\F}}{\|\Phi^\circ\|_{\F}} \nn} 
	and is real-valued.
	Fig.~\ref{fig:ave_ptw_rel_err_MC} shows the relative error averaged over the $100$ trials, and	
	Fig.~\ref{fig:int_rel_err_MC} shows the boxplot of {the integrated relative error} for each one of the four spectral estimators.	
	We can clearly see that VARMA outperform the other estimators. {In particular, it is interesting to note that although VAR-ma estimates the spectral density of an ARMA process, its inferior performance is due to the fact that the latter searches the process over a special class of ARMA processes, as pointed out in Remark \ref{RemSCAL}.}

We have also tested the impact of $U^{(0)}$, the initial inner function for the alternating minimization scheme \eqref{step1} and \eqref{step2}. More precisely, we fix a sample path of length $500$ of the true model in Subsec.~\ref{subsec:para_set}. Then we do another Monte Carlo simulations of $100$ trials to compute the VARMA estimate, and each trial starts from a different $U^{(0)}$ whose value is a constant orthogonal matrix for all $\vartheta\in\Tbb$. Such an orthogonal matrix is generated by applying the QR factorization to a random square matrix with i.i.d.~standard normal entries \cite{diaconis2003patterns}.
In the end, we compute the difference between two estimators $\hat{\Phi}_{j}$ and $\hat{\Phi}_{k}$ where the subscripts indicate the index of a trial. We find that	$\int\|\hat{\Phi}_{j}-\hat{\Phi}_{k}\|_{\F} \approx 10^{-4}$ for all $j\neq k$, which means that all resulting spectral estimates are essentially the same.


\section{Conclusions}\label{sec:conclus}

We have developed a new paradigm for spectral estimation of {vector-valued} ARMA (VARMA) processes based on the philosophy of Maximum Entropy. A new transportation entropy is employed as the objective function and a new concept of matricial cepstral coefficients, along with the covariances, is used in the equality constraints to select the spectrum. The optimization problem is treated via Lagrange dual analysis. We show that the dual problem can be made well-posed when a suitable regularization term is added. A spectral estimation procedure based on a finite sample path of {a random process has  also been proposed}. Numerical simulations confirm the advantage of our estimator when the true spectrum has a nontrivial zero structure.

For future work, it is possible to exploit the subspace inclusion condition \eqref{cond3} or \eqref{cond4} via techniques in semidefinite programming in order to obtain an estimation theory for low-rank spectral densities \cite{cao2023identification,cao2023modeling}.

\bibliographystyle{IEEEtran}
\bibliography{references}

 \end{document}